\documentclass{article}

\usepackage{microtype}
\usepackage{graphicx}
\usepackage{subfigure}
\usepackage{booktabs} 

\usepackage[pagebackref,breaklinks,colorlinks]{hyperref}
\usepackage{caption}

\usepackage{algorithmic}
\newenvironment{mymodule}[1][htb]
  {
   \begin{algorithm}[#1]%
  }{\end{algorithm}}
\usepackage[algo2e]{algorithm2e} 
\newcommand{\D}[1]{$\mathcal{D}_{#1}$}

\newcommand{\N}[1]{$\mathcal{N}_{#1}$}

\usepackage[accepted]{icml/icml2023}

\usepackage{amsmath}
\usepackage{amssymb}
\usepackage{mathtools}
\usepackage{amsthm}

\usepackage[capitalize,noabbrev]{cleveref}

\theoremstyle{plain}

\theoremstyle{definition}

\theoremstyle{remark}

\usepackage[textsize=tiny]{todonotes}

\icmltitlerunning{CrossSplit: 
Mitigating  Label Noise  Memorization through Data Splitting}

\begin{document}

\twocolumn[
\icmltitle{CrossSplit: 
Mitigating  Label Noise  Memorization through Data Splitting
}



\icmlsetsymbol{equal}{*}

\begin{icmlauthorlist}
\icmlauthor{Jihye Kim}{sait,sail2}
\icmlauthor{Aristide Baratin}{sail}
\icmlauthor{Yan Zhang}{sail}
\icmlauthor{Simon Lacoste-Julien}{sail,mila,cifar}
\end{icmlauthorlist}

\icmlaffiliation{sait}{Samsung Advanced Institute of Technology (SAIT), Suwon, South Korea}
\icmlaffiliation{sail}{SAIT AI Lab, Montreal, Canada}
\icmlaffiliation{sail2}{Work done as a visiting researcher at SAIT AI Lab, Montreal, Canada}
\icmlaffiliation{mila}{Mila, Université de Montreal, Canada}
\icmlaffiliation{cifar}{Canada CIFAR AI Chair}

\icmlcorrespondingauthor{Jihye Kim}{jihye32.kim@samsung.com}

\icmlkeywords{label noise, cross-split label correction, data splitting}

\vskip 0.3in
]



\printAffiliationsAndNotice{}  

\begin{abstract}

We approach the problem of improving robustness of deep learning algorithms in the presence of label noise. Building upon existing label correction and co-teaching methods, we propose a novel training procedure to mitigate the memorization of noisy labels, called CrossSplit, which uses a pair of neural networks trained on two disjoint parts of the labelled dataset. CrossSplit combines two main ingredients: $(i)$ Cross-split label correction. The idea is that, since the model trained on one part of the data cannot memorize example-label pairs from the other part, the training labels presented to each network can be smoothly adjusted by using the predictions of its peer network; $(ii)$ Cross-split semi-supervised training. A network trained on one part of the data also uses the unlabeled inputs of the other part.  
Extensive experiments on CIFAR-10, CIFAR-100, Tiny-ImageNet and mini-WebVision datasets demonstrate that our method can outperform  the current state-of-the-art in a wide range of noise ratios.  
\end{abstract}

\section{Introduction}
\label{sec:intro}

\begin{figure*}[t]
  \centering
  \includegraphics[width=0.9\linewidth]{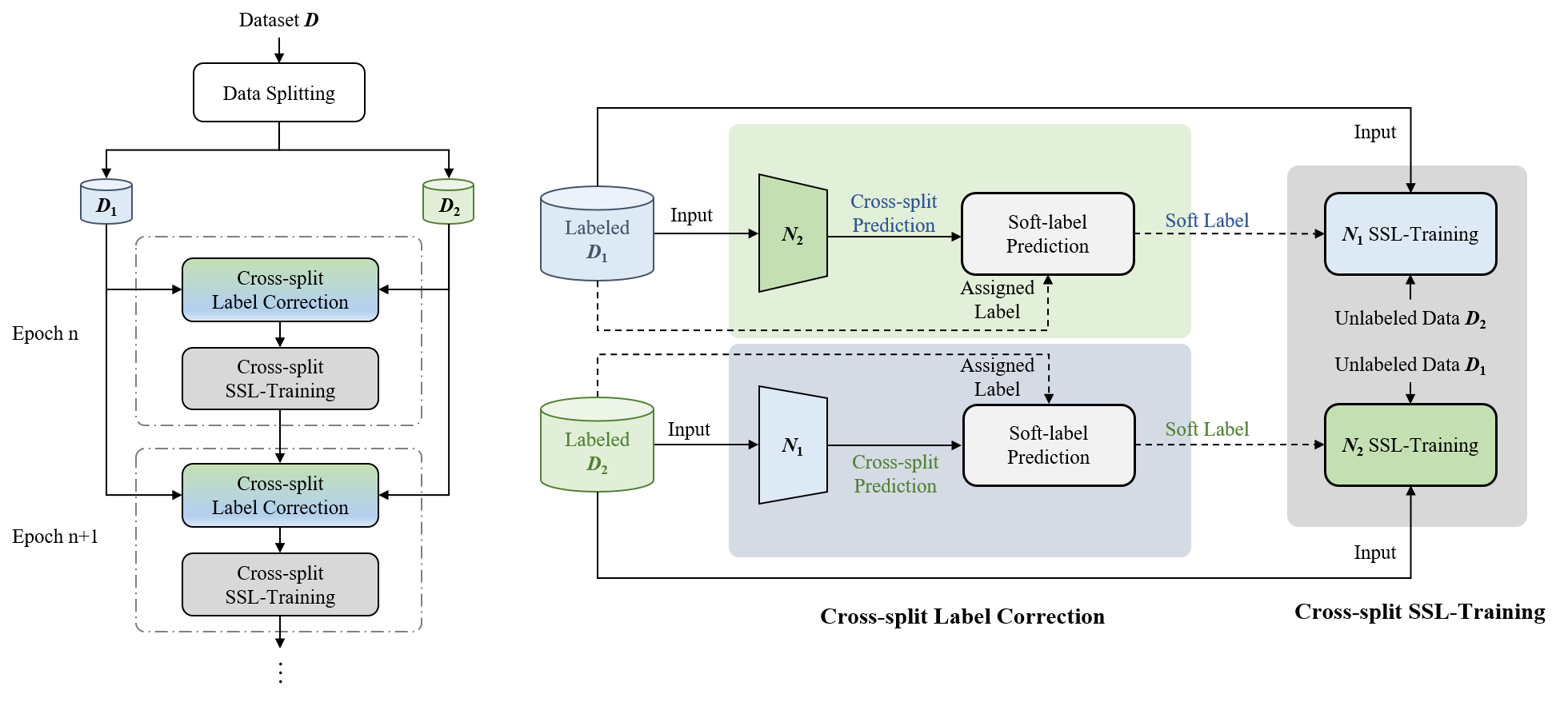}
  \caption{
\emph{CrossSplit} splits the original training labelled dataset into two disjoint parts and trains a separate  network on each of these splits. The dataset each network is trained on is also used by the peer network as unlabeled data  for semi-supervised learning (SSL). At each training epoch, \emph{CrossSplit} uses a cross-split label correction scheme that defines soft labels in terms of the peer prediction. 
  }
  \label{fig:Netarchitecture}
\end{figure*}

A large part of the success of deep learning algorithms relies on
the availability of massive amounts of labeled data, via e.g. web crawling \cite{li2017webvision} or crowd-sourcing platforms \cite{song2019selfie}.  
However, while these data-collection methods enable to bypass cost-prohibitive human annotations,  they 
inherently yield a lot of mislabeled samples \cite{xiao2015learning, li2017webvision}. 
This leads to a degradation of the performance, especially considering that deep neural  networks have enough capacity to fully memorize noisy labels \cite{zhang2017understanding, liu2020early, arpit2017closer}. An important issue in the field is therefore to adapt the training process to improve robustness under label noise.

This problem has been addressed in various ways in the recent literature. Two common approaches are {\it label correction} and {\it sample selection}. The first one focuses on correcting the noisy labels during training, e.g. by using soft labels defined as convex combinations of the assigned label and the model prediction \cite{reed2015training, arazo2019unsupervised, lu2022selc}. 
Another common approach uses sample selection mechanisms, which separate clean examples from noisy ones during training \cite{dividemix, karim2022unicon, han2018co, yu2019does}, e.g. using a small-loss criterion \cite{li2019learning}. Current state-of-the-art methods \cite{dividemix, karim2022unicon} combine epoch-wise sample selection with a  co-teaching procedure \cite{han2018co, yu2019does}
where two networks are trained simultaneously, each of them using the sample selection of the other so as to mitigate confirmation bias. Semi-supervised learning (SSL) techniques are then used where the selected noisy examples are treated as unlabeled data. 

Despite the popularity and success of these methods, they are not exempt from drawbacks. Existing label correction methods define soft target labels in terms of their own prediction, which may become unreliable as training progresses and memorization occurs \cite{lu2022selc}. Sample selection procedures rely on criteria to filter out noisy examples which are subject to selection errors 
-- 
in fact, making an accurate distinction between mislabelled and inherently difficult examples is a notoriously challenging problem \cite{Hooker_tail,pleiss2020identifying, baldock2021deep}. 

The goal of this paper is to propose a novel robust training scheme that addresses some of these drawbacks. The idea is to bypass the sample selection process by using a random splitting of the data into two disjoints parts, and to train a separate network on each of these splits. The rationale is that the model trained on one part of the data cannot memorize input-label pairs from the other part. We propose to correct the labels presented to each network by using a combination of the assigned label and the prediction of the peer network. This procedure  allows us to avoid the memorization of examples without significantly degrading the learning of difficult examples. Cross-split semi-supervised learning is then performed where 
the data each network is trained on is also used as unlabeled data by the peer network.

\begin{figure*}[t]
  \centering
  \includegraphics[width=1.\linewidth]{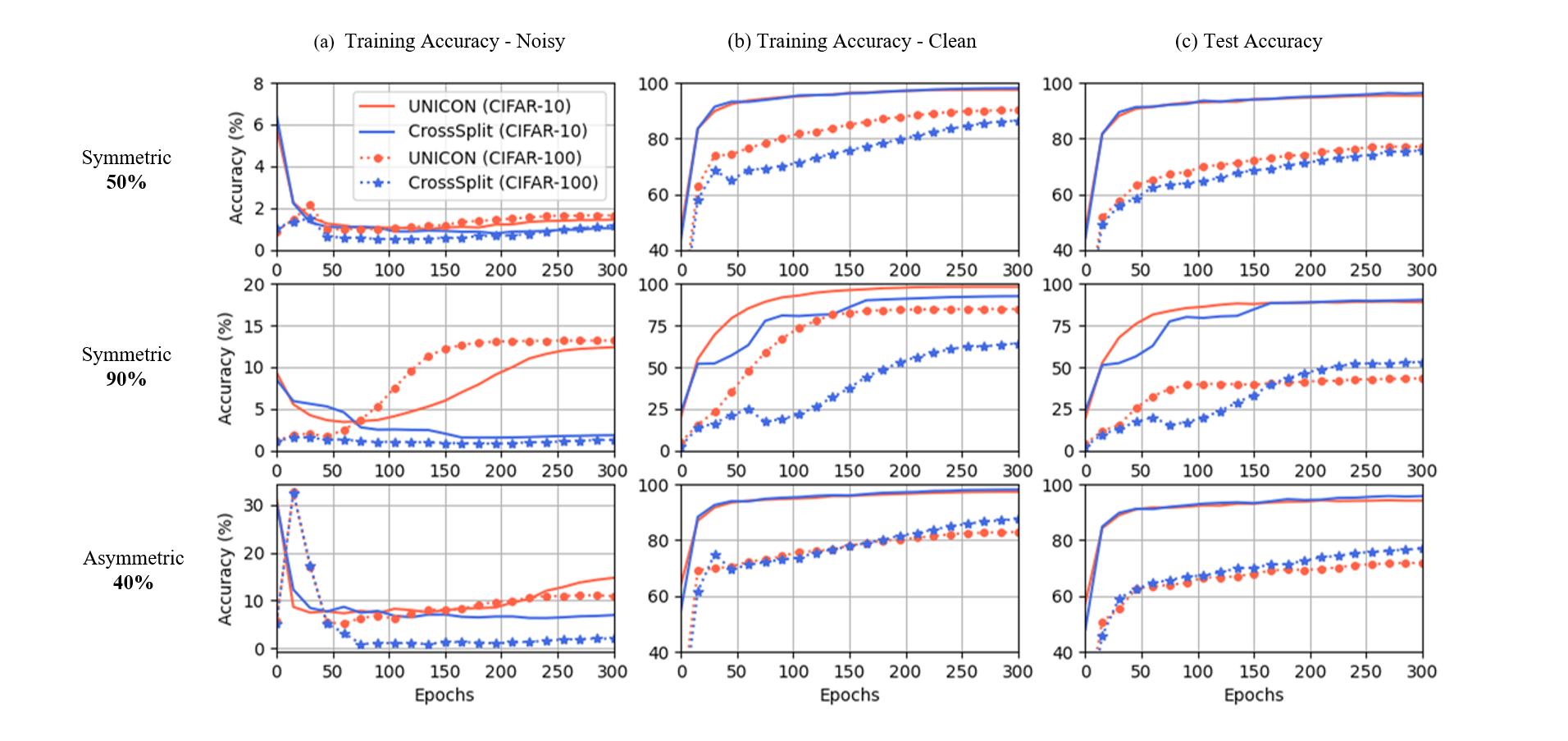}

   \caption{Memorization of clean and noisy training samples of CIFAR-10 and CIFAR-100 for different types of noise and noise ratio. Compared to UNICON \cite{karim2022unicon}, \emph{CrossSplit} induces less memorization (lower accuracy) on the noisy labels while having comparable accuracy on clean samples. It is interesting to note that in the case of a very high noise ratio (90$\%$), \emph{CrossSplit} has a lower training accuracy on clean data than UNICON, yet yields a higher test performance. This shows how important  reducing memorization is, since the lower memorization of noisy labels completely offsets the lower accuracy on clean samples.}
  \label{fig:memo}
  \vspace{-2mm}
\end{figure*}

Our contributions are summarized as follows:

\begin{itemize}
\item We introduce \emph{CrossSplit} for robust training (\cref{sec:method}, overview in \cref{fig:Netarchitecture}). CrossSplit departs from existing methods  
by using a pair of networks trained on \emph{two random splits} of the labeled dataset, leading to a novel \emph{label 
correction procedure} based on peer-predictions and a cross-split semi-supervised training process.  

\item Through experimental analysis, we verify that this data splitting and training scheme 
help in reducing the memorization of noisy labels (\cref{fig:memo}), which in turn improves robustness under label noise. 

\item Through extensive experiments on 
CIFAR-10, CIFAR-100, Tiny-ImageNet, and mini-WebVision datasets, we show that our method can outperform  the current state-of-the-art in a wide range of noise ratios (\cref{sec:exp}).

\item We perform a thorough ablation study of the different components of our procedure (\cref{sec:ablation}).
\end{itemize}

\section{Proposed Method}
\label{sec:method}

In this section we introduce \emph{CrossSplit} 
for alleviating memorization of noisy labels in order to improve robustness.

{
\setlength{\textfloatsep}{0.1cm}
\setlength{\floatsep}{0.1cm}
\begin{mymodule}[t]
\caption{CrossSplit: Cross-split SSL training based on cross-split label correction}
\textbf{Input:} Split training set $\mathcal{D}={\{\mathcal{D}_1,\mathcal{D}_2\}}$, pair of networks \N1, \N2, warmup epoch $E_{\text{warm}}$, total number of epochs $E_\text{max}$.\\
$\theta_1$, $\theta_2 \, \leftarrow \, $ Initialize network parameters \\
$\theta_1$, $\theta_2 \, \leftarrow \, $ Warmup supervised training on whole dataset for $E_{\text{warm}}$ epochs

\For{epoch $\in [E_{\text{warm}}+1,\hdots,E_\text{max}]$}
{
1. Training \N1: \\
1.1: Perform cross-split label correction \cref{eq:softlabel} for labeled $\mathcal{D}_1$ using the predictions of \N2 (see \cref{subsec:label_corr}).

1.2: Perform SSL training \cite{sohn2020fixmatch} using (soft)-labeled \D1 as labeled data and \D2 as unlabeled data (see \cref{subsec:cstraining}). \\
2. Analogous training for \N2.
}

\textbf{Return:} $\theta_1, \theta_2$.
\label{mo:pseudocode}
\end{mymodule}
\vspace{-4mm}
}

\paragraph{Setup} Just like in standard co-training \cite{blum1998combining} and co-teaching  \cite{han2018co, yu2019does} schemes, \emph{CrossSplit} simultaneously trains two neural networks 
\N1 and \N2. While these networks can in principle be completely different models, for simplicity we use the same architecture with two distinct sets of parameters. Our procedure begins with a random splitting of the labeled dataset 
\D{} into two disjoint subsets \D1 and \D2 of equal size. At each training epoch, \emph{CrossSplit} includes a label correction step where the labels presented to each network are corrected  using the peer network prediction.
This is a simple yet effective way to mitigate memorization of the noisy labels, since each network cannot memorize the input-label pairs presented to its peer. Following \cite{dividemix,karim2022unicon}, \emph{CrossSplit} then leverages  semi-supervised learning  techniques; 
the novelty here is to bypass the  usual sample selection of noisy data, and to rely instead on a mere cross-split training:  \N1 is trained on \D1 (with soft labels) and uses the inputs of \D2 as unlabeled data;  \N2 is trained on \D2 (with soft labels) and uses the inputs of \D1 as unlabeled data. The training procedure is illustrated in \cref{fig:Netarchitecture} and summarized in \cref{mo:pseudocode}. 

We provide below a more detailed description of the different components of \emph{CrossSplit}.

\subsection{Cross-split Label Correction}
\label{subsec:label_corr}

Label correction serves the important purpose of identifying which examples are likely to be mislabeled. At every epoch of our training procedure, for each of the two networks, we will use soft labels defined as convex combinations of the assigned label and the peer network prediction. The crucial aspect is that due to the data splitting, the peer network cannot memorize the label that it is modifying. This is in contrast to existing methods \cite{reed2015training, dividemix, karim2022unicon, lu2022selc} that combine assigned labels with the network's own prediction: if the network has memorized the noisy label, it simply reinforces the mislabeling.

Consider the network \N1 and let $(\mathbf{x}_i, \mathbf{y}_i) \in \mathcal{D}_1$, where $\mathbf{x}_i$ is an input image and $\mathbf{y}_i$ is the one-hot vector associated to its (possibly noisy) class label.  We define the  soft label $\mathbf{s}_i$ 
as the following convex combination of $\mathbf{y}_i$ and the cross-split probability (softmax) vector, $\mathbf{\hat{y}}_{\text{peer},i} = \mathcal{N}_2(\mathbf{x}_i)$:
\begin{align}
\mathbf{s}_i &= {\beta_i \mathbf{\hat{y}}_{\text{peer},i} + (1 - \beta_i) \mathbf{y}_i}
\label{eq:softlabel}\\
{\beta}_i &= \gamma(\mathrm{JSD}_\text{norm}(\mathbf{\hat{y}}_{\text{peer},i}, \mathbf{y}_i) - 0.5) + 0.5
\label{eq:beta}
\end{align}
where $\mathrm{JSD}_\text{norm}$ is a normalized version of the Jensen-Shannon Divergence (JSD) described in \cref{eq:norm_of_JSD} below, and $\gamma$ is a relaxation parameter.\footnote{This parameter enables us to control the range of $\beta_i$, especially at the beginning of training where we may expect the JSD values to be noisy. 
We explain this in more detail in  \cref{relaxation}.}  
Intuitively, when the peer network confidently predicts the assigned label $\mathbf{y}_i$, ${\beta}_i$ is small and \cref{eq:softlabel} picks a soft label that is close to $\mathbf{y}_i$. For a confident peer prediction that disagrees with $\mathbf{y}_i$, the soft label shifts towards the cross-prediction label $\mathbf{\hat{y}}_{\text{peer},i}$.

\paragraph{Class-balancing coefficient normalization}
\label{subsec:CWN}
{UNICON} \cite{karim2022unicon} noted that when performing sample selection, the selection threshold should vary between different classes.
Otherwise, the model is biased towards selecting samples from easy classes to be clean, while rejecting clean samples from harder classes as noisy.
We can adapt this idea to our framework by thinking of the weighting from $\beta_i$ as ``soft'' sample selection.
In particular, we normalize the standard JSD that \citet{karim2022unicon} use in such a way  that, \emph{within each class}, it ranges from 0 to 1.

To compute this, we keep track of the minimum and maximum $\mathrm{JSD}$ values within each class, which we compute at the beginning of every epoch. 
For each class, encoded by the one-hot vector $\mathbf{y}$,  we thus compute the quantities 
\begin{align}
    \mathrm{JSD}^{\mathrm{min}}_{\mathbf{y}} & : = \min_{\{j | \mathbf{y}_j = \mathbf{y}\}} \mathrm{JSD}(\mathbf{\hat{y}}_{\text{peer},j}, \mathbf{y})
    \nonumber \\
    \mathrm{JSD}^{\mathrm{max}}_{\mathbf{y}} &:= \max_{\{j | \mathbf{y}_j = \mathbf{y}\}} \mathrm{JSD}(\mathbf{\hat{y}}_{\text{peer},j}, \mathbf{y})
    \label{eq:JSD_minmax}
\end{align}
For each example, we then normalize the JSD  through shifting and scaling,  using the values (\cref{eq:JSD_minmax}) associated to its class. 
\begin{equation}
    \mathrm{JSD}_{\text{norm}}(\mathbf{\hat{y}}_{\text{peer},i}, \mathbf{y}_i)  := 
\cfrac
    {\mathrm{JSD}(\mathbf{\hat{y}}_{\text{peer},i}, \mathbf{y}_i) - \mathrm{JSD}^{\mathrm{min}}_{\mathbf{y}_i} }
    {\mathrm{JSD}^{\mathrm{max}}_{\mathbf{y}_i} - \mathrm{JSD}^{\mathrm{min}}_{\mathbf{y}_i} }
    \label{eq:norm_of_JSD}
\end{equation}

\subsection{Cross-split SSL-Training}
\label{subsec:cstraining}
\N1 and \N2 are each trained on only half the amount of labeled data, which can degrade performance.
We thus look towards semi-supervised learning, which lets us train \N1 using \emph{unlabeled} data (to avoid memorization) from \D2 and \N2 using unlabeled data from \D1.

We use a cross-split semi-supervised training procedure. At each training epoch, \N1 is trained on (soft)-labeled \D1 with the unlabeled samples from \D2 and \N2 is trained on (soft)-labeled \D2 with the unlabeled samples from \D1 (see \cref{fig:Netarchitecture}). Regarding the specific techniques used, we reproduce the main ingredients of existing methods \cite{dividemix,karim2022unicon}, by following  FixMatch \cite{sohn2020fixmatch} and applying MixUp \cite{zhang2018mixup} augmentation.  Just like {UNICON} \cite{karim2022unicon}, the semi-supervised loss is combined with a contrastive loss evaluated on the unlabeled dataset to further mitigate noisy label memorization.

\section{Related Work}
\label{sec:relt} 
 \begin{figure}[t]
   \centering
   \includegraphics[width=1.\linewidth]{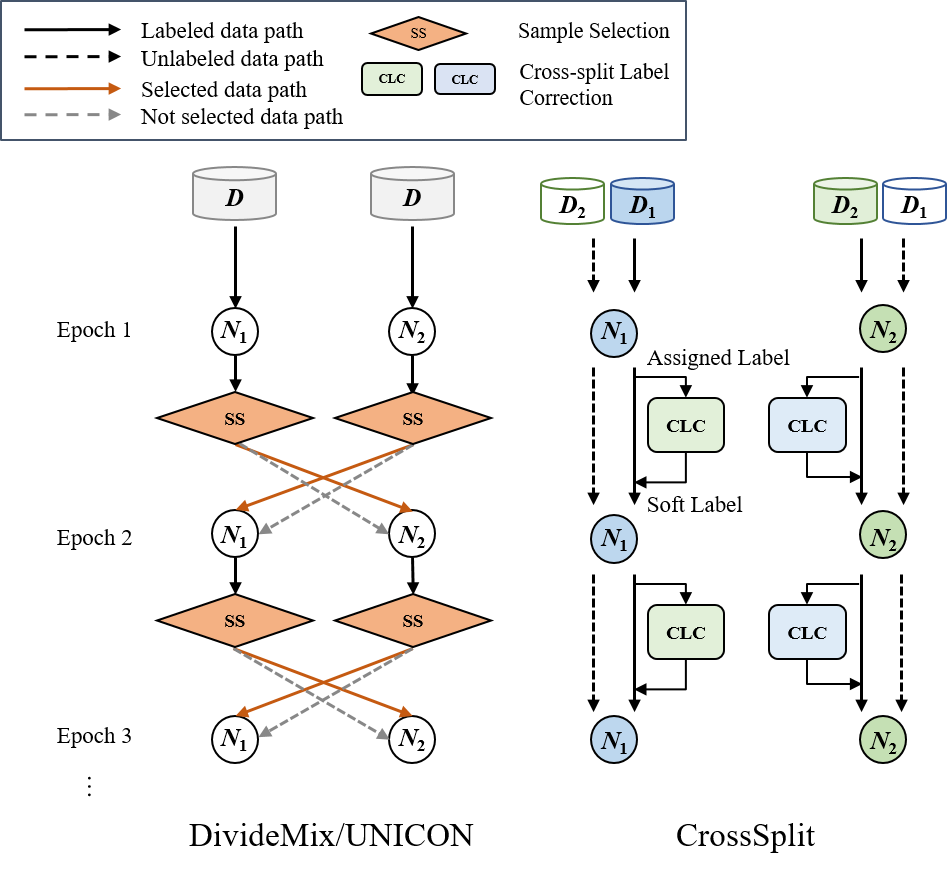}
   \vspace{-4mm}
   \caption{
   Comparison of the DivideMix \cite{dividemix}, UNICON \cite{karim2022unicon}, and \emph{CrossSplit} co-teaching pipelines. The data flow is represented with solid lines for labeled data and dotted lines for unlabeled data. All three methods train two networks (\N1 \& \N2) simultaneously. In DivideMix and UNICON, at every epoch, each network separates clean samples (orange solid line) and noisy samples (gray dotted line) using a small loss criterion, and transfers the two subsets to its peer network for subsequent semi-supervised learning. By contrast, \emph{CrossSplit} splits the original training dataset into two halves and trains each network on one of these splits. For each of the two networks, we use soft labels defined as convex combinations of the assigned label and the peer network prediction via cross-split label correction (CLC) process. The data each network is trained on is also used by the peer network as unlabeled data for semi-supervised learning. 
   }
   \label{fig:dataflow}
   \vspace{-2mm}
 \end{figure}

The problem of learning with noisy labels has been approached in various ways in the literature. These include label correction \cite{reed2015training, arazo2019unsupervised, zhang2020learning, dividemix, lu2022selc}, noise robust loss \cite{zhang2018generalized, ma2020normalized}, loss correction \cite{goldberger2017training} and sample selection \cite{dividemix, karim2022unicon, han2018co, yu2019does} based methods. Most relevant to our work are \emph{label correction} and \emph{sample selection}, which we discuss now in more detail.

\paragraph{Label correction methods} In order to mitigate the negative influence of noisy labels in training, some works have focused on gradually adjusting the assigned label based on the model's prediction \cite{reed2015training, arazo2019unsupervised, zhang2020learning, dividemix, lu2022selc, ma2018dimensionality, tanaka2018joint}.
{\it Bootstrapping} \cite{reed2015training} generates new regression targets 
by combining the assigned label and the model's prediction, using the same fixed combination weight for all samples. 
{\it M-correction} \cite{arazo2019unsupervised} uses instead dynamic weights defined in terms of the sample's training loss values. 
Follow-up works proposed to incorporate the prediction confidence or use ensemble predictions by an exponential moving average of network in the design of the weights \cite{zhang2020learning, lu2022selc}. However, existing label correction methods have a limitation in that labels are corrected only based on their own prediction.
This can lead to the memorization of noisy samples which only reinforces their own mislabelings.
This is in contrast to the label correction method proposed in our work, which generates soft labels as combinations  of the assigned label and the peer network prediction; the peer network cannot memorize the label because it never sees that label during its own training.

\paragraph{Sample selection-based methods} Another  common approach is to identify the noisy samples, e.g., using a small-loss criterion \cite{li2019learning}, to separate them from the clean ones, and to use the two subsets of samples in a different way during training \cite{han2018co, dividemix, karim2022unicon}. The selected clean set is typically used for conventional supervised learning; the noisy samples are either excluded from training \cite{han2018co} or treated as unlabeled data for semi-supervised learning \cite{dividemix, karim2022unicon}.

Despite the empirical success of this approach, it has some limitations. Sample selection processes  require hyper-parameter setting for selection (e.g., noise ratio, threshold value for selection); how well this selection is done affects performance. They also face the difficult challenge to distinguish  between mislabeled data, which should not be memorized,  and difficult examples whose labels nevertheless carry useful information \cite{Feldman2019}. Our work proposes a method that bypasses the sample selection process, where 
the hard decision as to whether a sample is clean or not is replaced by a soft label correction using the peer network. 

\paragraph{Co-training methods} State-of-the-art sample selection methods take advantage of training two models simultaneously in order to prevent confirmation bias \cite{dividemix, karim2022unicon}. One network selects its small-loss samples (considered as clean samples) to teach its peer network for subsequent training \cite{han2018co,yu2019does}. This idea of network cooperation can be traced back to co-training \cite{blum1998combining}, which can be shown to improve the performance of learning by unlabeled data in semi-supervised learning.  In the original version  \cite{blum1998combining}, multiple classifiers are trained on distinct views of the data, e.g., mutually exclusive feature sets for the same example, and exchange their predictions.
For example, data with high confidence prediction can be added for re-training to the data seen by the peer model \cite{ma2017self}.
Our approach is similar in spirit, but instead of working with different views of the data, we train different models on disjoint subsets of the dataset.
\cref{fig:dataflow} illustrates the differences between \emph{CrossSplit} and several other co-teaching schemes used in the recent literature \cite{han2018co, dividemix, karim2022unicon}.

\section{Experiments}
\label{sec:exp}

\subsection{Datasets}

We conduct experiments both on datasets with simulated label noise and datasets with natural label noise.  Simulating the noise allows us to control the noise level, analyze the memorization behavior of our algorithm and test a variety of scenarios. On the other hand, working with naturally noisy datasets enables  practical evaluation in situations where the type and level of noise are unknown.

The CIFAR-10/100 datasets \cite{krizhevsky2009learning} each contain 50K training  and 10K testing 32 $\times$ 32 coloured images. Following the setup of previous works \cite{Li2017LearningFN, Tanaka2018JointOF, yu2019does, dividemix, karim2022unicon}, we use both symmetric and asymmetric label noise. Symmetric label noise is generated by re-assigning to a portion of the training data in each class, a label chosen uniformly at random among all other classes. Asymmetric label noise mimics real-world label noise more closely: the labels are chosen among similar classes (e.g.,  Bird $\xrightarrow[]{}$ Airplane, Deer $\xrightarrow[]{}$ Horse, Cat $\xrightarrow[]{}$ Dog). For CIFAR-100, labels are flipped circularly within the super-classes.
We simulate a wide range of noise levels: 20\% - 90\% for symmetric label noise and 10\% - 40\% for asymmetric label noise.

\paragraph{Tiny-ImageNet} \cite{le2015tiny} is a subset of the ImageNet dataset with 100K  64 $\times$ 64 coloured images distributed within 200 classes. 
Each class has 500 training images, 50 test images and 50 validation images.
We experiment on Tiny-ImageNet with simulated symmetric label noise.

\paragraph{mini-WebVision} \cite{li2017webvision} contains 2.4 million images from websites Google and Flicker and contains many naturally noisy labels. The images are categorized into 1,000 classes and following \citet{karim2022unicon}, we use the top-50 classes from the Google images of WebVision for training.

\subsection{Experimental details} 

\paragraph{Architectures} 
For CIFAR-10, CIFAR-100 and Tiny-ImageNet, in line with \cite{dividemix, karim2022unicon}, we use a PreAct ResNet18 \cite{he2016deep} architecture. 
For mini-WebVision, following \cite{ortego2021multi}, we use ResNet18. 
We give training details in \cref{training_details}.

\subsection{Results}

In this section, we compare  the performance of \emph{CrossSplit} with existing methods (\cref{sec:results_perf}), which include label correction and sample-selection methods. We also analyze the memorization behaviour of the algorithm (\cref{sec:results_memo}). Our baselines are Bootstrapping \cite{reed2015training}, JPL \cite{kim2021joint}, M-Correction \cite{arazo2019unsupervised}, MOIT \cite{ortego2021multi}, SELC \cite{lu2022selc}, Sel-CL\cite{li2022selective}, DivideMix \cite{dividemix}, ELR \cite{liu2020early}, and UNICON \cite{karim2022unicon}.

\begin{table*}[t]

\small
{Tables 1 \& 2. Test accuracy (\%) comparison on CIFAR-10 (left) and CIFAR-100 (right)  with symmetric and asymmetric label noise.
Our model achieves state-of-the-art performance on almost every dataset-noise combination.
The best scores are {\bf boldfaced}, and the second best ones are \underline{underlined}. The baseline results are imported from \cite{karim2022unicon, dividemix,li2022selective} and sorted according to their performance in the case of a 20\% symmetric noise ratio.}
\vspace{-2mm}

\begin{minipage}[b]{0.5\textwidth}

\centering

\caption{CIFAR-10}
\label{tab:cifar10}

{
\renewcommand{\arraystretch}{1.1}
\setlength\tabcolsep{8pt}
\renewcommand{\tabcolsep}{1.0 mm}

\resizebox{\linewidth}{!}{
\begin{tabular}{c| c|c c c c|c c c }

\toprule

\multicolumn{2}{c|}{{Noise type}} & \multicolumn{4}{c|}{{Symmetric}} & \multicolumn{3}{c}{{Asymmetric}}\\
\multicolumn{2}{c|}{Method/Noise ratio} & $20\%$ & $50\%$ & $80\%$ & $90\%$  & $10\%$ & $30\%$ & $40\%$\\
\midrule
\multicolumn{2}{c|}{CE}
& 86.8 & 79.4 & 62.9 & 42.7 & 88.8 & 81.7 & 76.1 \\
\multicolumn{2}{c|}{Bootstrapping ~\cite{reed2015training}} %
& 86.8 & 79.8 & 63.3 & 42.9 & - & -& - \\ 
\multicolumn{2}{c|}{JPL ~\cite{kim2021joint}} 
& 93.5 & 90.2 & 35.7 & 23.4 & 94.2 & 92.5 & 90.7\\
\multicolumn{2}{c|}{M-Correction ~\cite{arazo2019unsupervised}}
& 94.0 & 92.0 & 86.8 & 69.1 & 89.6 & 92.2 & 91.2  \\
\multicolumn{2}{c|}{MOIT ~\cite{ortego2021multi}} 
& 94.1 & 91.1 & 75.8 & 70.1 & 94.2 & 94.1 & 93.2\\
\multicolumn{2}{c|}{SELC ~\cite{lu2022selc}} 
& 95.0 & - & 78.6 & - & - & - & 92.9\\
\multicolumn{2}{c|}{Sel-CL ~\cite{li2022selective}} 
& 95.5 & 93.9 & 89.2 & {81.9} &\underline{95.6} & \underline{95.2} & 93.4\\
\multicolumn{2}{c|}{MixUp ~\cite{zhang2018mixup}}
& 95.6 & 87.1 & 71.6 & 52.2 & 93.3 & 83.3 & 77.7 \\
\multicolumn{2}{c|}{ELR ~\cite{liu2020early}} 
& 95.8 & 94.8 & 93.3 & {78.7} & 95.4 & 94.7 & 93.0\\
\multicolumn{2}{c|}{UNICON ~\cite{karim2022unicon}} 
& 96.0 & \underline{95.6} & \underline{93.9} & \underline{90.8} & 95.3 & 94.8 & \underline{94.1} \\
\multicolumn{2}{c|}{DivideMix ~\cite{dividemix}}
& \underline{96.1} & 94.6 & 93.2 & 76.0 & 93.8 & 92.5 & 91.7 \\

\midrule

\multicolumn{2}{c|}{CrossSplit (ours)} & \textbf{96.9} & \textbf{96.3} & \textbf{95.4} & \textbf{91.3} &  \textbf{96.9} & \textbf{96.4} & \textbf{96.0}\\

\bottomrule

\end{tabular}
}
}

\end{minipage}
\begin{minipage}[b]{0.50\textwidth}
\centering
\caption{CIFAR-100}
\label{tab:cifar100}
\small
{
\renewcommand{\arraystretch}{1.1}
\setlength\tabcolsep{8pt}
\renewcommand{\tabcolsep}{1.0 mm}
\resizebox{\linewidth}{!}{
\begin{tabular}{c|c|c c c c|c c c}

\toprule

\multicolumn{2}{c|}{{Noise type}} & \multicolumn{4}{c|}{{Symmetric}} & \multicolumn{3}{c}{{Asymmetric}}\\
\multicolumn{2}{c|}{Method/Noise ratio} & $20\%$ & $50\%$ & $80\%$ & $90\%$  & $10\%$ & $30\%$ & $40\%$\\
\midrule
\multicolumn{2}{c|}{CE}
& 62.0 & 46.7 & 19.9 & 10.1 & 68.1 & 53.3 & 44.5\\
\multicolumn{2}{c|}{Bootstrapping ~\cite{reed2015training}} %
& 62.1 & 46.6 & 19.9 & 10.2 & - & -& - \\ 
\multicolumn{2}{c|}{MixUp ~\cite{zhang2018mixup}}
& 67.8 & 57.3 & 30.8 & 14.6 & 72.4 & 57.6 & 48.1\\
\multicolumn{2}{c|}{JPL ~\cite{kim2021joint}} 
& 70.9 & 67.7 & 17.8 & 12.8 & 72.0 & 68.1 & 59.5\\

\multicolumn{2}{c|}{M-Correction ~\cite{arazo2019unsupervised}}
& 73.9 & 66.1 & 48.2 & 24.3 & 67.1 & 58.6 & 47.4 \\

\multicolumn{2}{c|}{MOIT ~\cite{ortego2021multi}} 
&  75.9 & 70.1 & 51.4 & 24.5 & 77.4 & 75.1 & 74.0\\
\multicolumn{2}{c|}{SELC ~\cite{lu2022selc}} 
& 76.4 & - & 37.2 & - & - & - & 73.6\\
\multicolumn{2}{c|}{Sel-CL ~\cite{li2022selective}} 
& 76.5 & 72.4 & 59.6 & \underline{48.8} & \underline{78.7} & \underline{76.4} & 74.2\\
\multicolumn{2}{c|}{DivideMix ~\cite{dividemix}}
&  77.3 & 74.6 & 60.2 & 31.5 & 71.6 & 69.5 & 55.1\\

\multicolumn{2}{c|}{ELR ~\cite{liu2020early}} 
& 77.6 & 73.6 & 60.8 & 33.4 & 77.3 & 74.6 & 73.2\\

\multicolumn{2}{c|}{UNICON ~\cite{karim2022unicon}} 
& \underline{78.9} & \textbf{77.6} & \underline{63.9} & 44.8& 78.2 & 75.6 & \underline{74.8}\\

\midrule

\multicolumn{2}{c|}{CrossSplit (ours)} & \textbf{79.9} &  \underline{75.7} &  \textbf{64.6} & \textbf{52.4}  & \textbf{80.7}& \textbf{78.5} &  \textbf{76.8}\\

\bottomrule
\end{tabular}
}
}

\end{minipage}
\end{table*}

\begin{table*}[t]
\small

{Tables 3 \& 4. Test accuracy (\%) comparison on Tiny-ImageNet (left) and Mini-WebVision (right).
Our model is competitive with the state-of-the-art (only small differences in performance) on Tiny-ImageNet with artificial noise, and surpasses the state-of-the-art on Mini-Webvision with real-world noise.
The best scores are {\bf boldfaced}, and the second best ones are \underline{underlined}. In Table 3, Best and Avg. mean  highest and average accuracy over the last 10 epochs. The baseline results  are imported from \cite{karim2022unicon} and  sorted according to their best performance in the case of a 20\% noise ratio. In Table 4, the baseline results are sorted by best performance.}
\vspace{-2mm}

\begin{minipage}{0.50\textwidth}

\centering
{
\renewcommand{\arraystretch}{1.0}
\setlength\tabcolsep{8pt}
\renewcommand{\tabcolsep}{1.0 mm}
\caption{Tiny-ImageNet}
\label{tab:tiny}
\resizebox{\linewidth}{!}{
\begin{tabular}{c| c|c c | c c}

\toprule

\multicolumn{2}{c|}{Noise type} & \multicolumn{4}{c}{Symmetric}  \\

\multicolumn{2}{c|}{Noise ratio} & \multicolumn{2}{c|}{$20\%$} & \multicolumn{2}{c}{$50\%$} \\
\midrule
\multicolumn{2}{c|}{Method} &  \multicolumn{1}{c}{Best} & \multicolumn{1}{c|}{Avg.} & \multicolumn{1}{c}{Best} & \multicolumn{1}{c}{Avg.} \\
\midrule
\multicolumn{2}{c|}{CE}
&  35.8 & 35.6 & 19.8 & 19.6 \\
\multicolumn{2}{c|}{Decoupling ~\cite{decoupling}}
&  37.0 & 36.3 & 22.8 & 22.6 \\
\multicolumn{2}{c|}{MentorNet ~\cite{jiang2018mentornet}} 
&  45.7 & 45.5 & 35.8 & 35.5\\
\multicolumn{2}{c|}{Co-teaching+ ~\cite{yu2019does}} 
&  48.2 & 47.7 & 41.8 & 41.2\\
\multicolumn{2}{c|}{M-Correction ~\cite{arazo2019unsupervised}}
&  57.2 & 56.6 & 51.6 & 51.3 \\
\multicolumn{2}{c|}{NCT \cite{sarfraz2021noisy}}
& 58.0 & 57.2 & 47.8 & 47.4 \\
\multicolumn{2}{c|}{UNICON ~\cite{karim2022unicon}} 
& \textbf{59.2} & \underline{58.4} & \textbf{52.7} & \textbf{52.4} \\ 
\midrule

\multicolumn{2}{c|}{CrossSplit (ours)} & \underline{59.1} & \textbf{58.8} & \underline{52.4} & \underline{52.0}\\ 
\bottomrule

\end{tabular}
}
}
\end{minipage}
\begin{minipage}{0.50\textwidth}
\centering
{
\renewcommand{\arraystretch}{1.0}
\setlength\tabcolsep{8pt}
\renewcommand{\tabcolsep}{1.0 mm}
\caption{Mini-WebVision}
\label{tab:webvision}
\resizebox{0.9\linewidth}{!}{
\begin{tabular}{c c| c c  }
\toprule

\multicolumn{2}{c|}{Method} & \multicolumn{1}{c}{Best}  & \multicolumn{1}{c}{Last}\\
\midrule
\multicolumn{2}{c|}{Decoupling ~\cite{decoupling}} & 62.54 & - \\
\multicolumn{2}{c|}{MentorNet ~\cite{jiang2018mentornet}} & 63.00 & - \\
\multicolumn{2}{c|}{Co-teaching ~\cite{han2018co}} & 63.58 & - \\
\multicolumn{2}{c|}{Iterative-CV ~\cite{chen2019understanding}} & 65.24 & - \\
\multicolumn{2}{c|}{ELR ~\cite{liu2020early}} & 73.00& 71.88 \\
\multicolumn{2}{c|}{SELC ~\cite{lu2022selc}} & 74.38 & - \\
\multicolumn{2}{c|}{MixUp ~\cite{zhang2018mixup}} & 74.96 & 73.76\\
\multicolumn{2}{c|}{DivideMix ~\cite{dividemix}} & 76.08 & \underline{74.64}\\
\multicolumn{2}{c|}{UNICON\cite{karim2022unicon}} & \underline{77.60} & - \\
\midrule
\multicolumn{2}{c|}{CrossSplit (ours)} & \bf{78.48} & \bf{78.07}\\

\bottomrule
\end{tabular}
}
}
\end{minipage}


\end{table*}

\subsubsection{Performance}
\label{sec:results_perf}

\cref{tab:cifar10} and \cref{tab:cifar100} show test accuracies on CIFAR-10 and CIFAR-100  with different levels of noise ratios ranging from 20\% to 90\% for symmetric noise and 10\% to 40\% for asymmetric noise respectively. 
We observe that \emph{CrossSplit} consistently outperforms the competing baselines  under a wide range of noise levels for the two types of noise models.
In particular, we note a large performance improvement in the case of asymmetric label noise (which is more likely to occur in real scenarios) for both CIFAR-10 and CIFAR-100. 
Even for symmetric label noise, we see performance improvements in all cases except for CIFAR-100 with a 50\% noise ratio. 
Additionally, we show visual comparisons of the features learned by UNICON \cite{karim2022unicon} and \emph{CrossSplit} in \cref{sec:tsne}. 
These show that the representations learned by our model are more distinct between classes, particularly when the noise is high.

For the Tiny-ImageNet dataset, we model symmetric label noise with two noise ratios, 20\% and 50\%. 
\cref{tab:tiny} shows test results both with the highest (Best) and the average over the last 10 epochs (Avg.).
In this case, compared to  existing algorithms, we observe a slight degradation of performance for a 50\% noise ratio with respect to the best competing baseline, and similar performance for a 20\% noise ratio.
Our results here are largely similar to the state-of-the-art.

\cref{tab:webvision} show performance comparisons for the mini-WebVision dataset, which is the most realistic task setting because the noise is present naturally due to the web-crawled nature of the data; the noise levels and structure of the noise are unknown. There is a 0.88 \% improvement over the current state-of-the-art UNICON \cite{karim2022unicon}, which demonstrates the benefits of our model in this experiment setting closest to the real world.

\begin{table*}[t]

\centering
\caption{{{\bf Ablation study on CIFAR-10}: Test accuracy (\%) of different setting on CIFAR-10 with varying noise rates (50\% - 90\% for Symmetric and 10\% - 40\% for Asymmetric noise).
We see that there is a minor difference when removing class-balancing normalization with lower noise ratios, but a large degradation in performance if it is removed for high noise ratios.
Mean and standard deviation of best and average of last 10 epochs are calculated over 3 repetitions of the experiments. The best results are highlighted in {\bf boldfaced} and scores that differ from them by more than 5\% are marked in 
{\color{red}red}.}} 
\label{tab:ablation_cifar10}

\small
{

\renewcommand{\arraystretch}{1.0}
\setlength\tabcolsep{8pt}
\renewcommand{\tabcolsep}{1.0 mm}
\begin{tabular}{c| c c cc cccc}
\toprule

\multicolumn{1}{c|}{Noise type} &\multicolumn{4}{c|}{Symmetric} &\multicolumn{4}{c}{Asymmetric} \\

\midrule

\multicolumn{1}{c|}{Noise ratio} & \multicolumn{2}{c|}{$50\%$}  & \multicolumn{2}{c|}{$90\%$} & \multicolumn{2}{c|}{$10\%$} & \multicolumn{2}{c}{$40\%$} \\
\midrule

\multicolumn{1}{c|}{Method} & \multicolumn{1}{c}{Best} & \multicolumn{1}{c|}{Last} & \multicolumn{1}{c}{Best} & \multicolumn{1}{c|}{Last} &\multicolumn{1}{c}{Best} & \multicolumn{1}{c|}{Last} &\multicolumn{1}{c}{Best} & \multicolumn{1}{c}{Last}   \\

\midrule

\multicolumn{1}{l|}{CrossSplit} &{96.34}\tiny${\pm0.05}$& \multicolumn{1}{c|}{{96.23}\tiny${\pm0.07}$}  & 
\bf{91.25}\tiny$\pm{0.79}$& \multicolumn{1}{c|}{{\bf91.02}\tiny$\pm{0.77}$} & \multicolumn{1}{c}{96.85\tiny$\pm0.04$} &\multicolumn{1}{c|}{96.74\tiny$\pm0.07$}&
\multicolumn{1}{c}{96.01\tiny$\pm0.12$}& \multicolumn{1}{c}{95.88\tiny$\pm0.13$}\\

\multicolumn{1}{l|}{w/o data splitting}&
96.10\tiny$\pm0.04$& \multicolumn{1}{c|}{95.96\tiny$\pm0.00$} &
90.30\tiny$\pm0.13$& \multicolumn{1}{c|}{89.93\tiny$\pm0.24$} &
96.76\tiny$\pm0.05$& \multicolumn{1}{c|}{96.63\tiny$\pm0.06$} &
92.16\tiny$\pm0.09$& \multicolumn{1}{c}{{\color{red}86.24}\tiny$\pm0.37$}\\

\multicolumn{1}{l|}{w/o class-balancing normalization} &
{\bf 96.73}\tiny$\pm0.13$ & \multicolumn{1}{c|}{{\bf 96.61}\tiny$\pm0.07$} & 
{\color{red}75.54}\tiny$\pm2.82$ & \multicolumn{1}{c|}{{\color{red}74.88}\tiny$\pm2.50$} & 
\bf{97.33}\tiny$\pm0.02$& \multicolumn{1}{c|}{{\bf97.20}\tiny$\pm0.02$} & 
{\bf 96.22}\tiny$\pm0.07$& \multicolumn{1}{c}{{\bf 96.04}\tiny$\pm0.12$}\\ 

\multicolumn{1}{l|}{w/o cross-split label correction} &
96.12\tiny$\pm0.05$& \multicolumn{1}{c|}{95.99\tiny$\pm0.03$} &
90.83\tiny$\pm0.25$& \multicolumn{1}{c|}{90.08\tiny$\pm0.40$} &
\bf{97.33}\tiny$\pm0.08$&\multicolumn{1}{c|}{{97.15}\tiny$\pm0.09$}&
96.12\tiny$\pm0.14$ & \multicolumn{1}{c}{95.95\tiny$\pm0.10$}\\

\bottomrule
\end{tabular}
}
\vspace{-2mm}
\end{table*}
\begin{table*}[t]

\centering
\caption{{{\bf Ablation study on CIFAR-100}: Test accuracy (\%) of different settings on CIFAR-100 with varying noise rates (50\% - 90\% for Symmetric and 10\% - 40\% for Asymmetric noise). With its higher difficulty than CIFAR-10, each component of \emph{CrossSplit} is crucial when the noise ratios are high.
Mean and standard deviation of best and average of last 10 epochs are calculated over 3 repetitions of the experiments. The best results are highlighted in {\bf boldfaced} and scores that differ from them by more than 5\% are marked in 
{\color{red}red}.}} 
\label{tab:ablation_cifar100}
\small
{

\renewcommand{\arraystretch}{1.0}
\setlength\tabcolsep{8pt}
\renewcommand{\tabcolsep}{1.0 mm}
\begin{tabular}{c| c c cc cccc}
\toprule


\multicolumn{1}{c|}{Noise type} &\multicolumn{4}{c|}{Symmetric} &\multicolumn{4}{c}{Asymmetric} \\

\midrule

\multicolumn{1}{c|}{Noise ratio} & \multicolumn{2}{c|}{$50\%$}  & \multicolumn{2}{c|}{$90\%$} & \multicolumn{2}{c|}{$10\%$} & \multicolumn{2}{c}{$40\%$} \\
\midrule
\multicolumn{1}{c|}{Method} & \multicolumn{1}{c}{Best} & \multicolumn{1}{c|}{Last} & \multicolumn{1}{c}{Best} & \multicolumn{1}{c|}{Last} &\multicolumn{1}{c}{Best} & \multicolumn{1}{c|}{Last} &\multicolumn{1}{c}{Best} & \multicolumn{1}{c}{Last}   \\

\midrule

\multicolumn{1}{l|}{CrossSplit} &{75.72}\tiny$\pm{0.18}$& \multicolumn{1}{c|}{{75.50}\tiny$\pm{0.18}$}  & 
\bf{52.40}\tiny$\pm{1.78}$& \multicolumn{1}{c|}{\bf{52.05}\tiny$\pm{1.94}$} & \multicolumn{1}{c}{80.71\tiny$\pm0.05$} & \multicolumn{1}{c|}{80.50\tiny$\pm0.06$}&\multicolumn{1}{c}{\bf{76.78}\tiny$\pm0.66$}& \multicolumn{1}{c}{\bf{76.56}\tiny$\pm0.55$}\\

\multicolumn{1}{l|}{w/o data splitting}&
73.63\tiny$\pm0.18$& \multicolumn{1}{c|}{73.36\tiny$\pm0.14$} &
{\color{red}14.19}\tiny$\pm1.30$& \multicolumn{1}{c|}{{\color{red}13.28}\tiny$\pm2.21$} &
78.97\tiny$\pm0.07$& \multicolumn{1}{c|}{78.77\tiny$\pm0.43$} &
72.12\tiny$\pm0.43$& \multicolumn{1}{c}{71.83\tiny$\pm0.42$}\\

\multicolumn{1}{l|}{w/o class-balancing normalization} &
{\bf 77.67}\tiny$\pm0.03$& \multicolumn{1}{c|}{{\bf 77.17}\tiny$\pm0.17$} & 
{\color{red}33.37}\tiny$\pm0.52$& \multicolumn{1}{c|}{{\color{red}18.53}\tiny$\pm0.19$} & 
{\bf 82.86}\tiny$\pm0.14$& \multicolumn{1}{c|}{{\bf 82.57}\tiny$\pm0.18$} & 
{\color{red}71.59}\tiny$\pm0.28$& \multicolumn{1}{c}{{\color{red}60.35}\tiny$\pm0.37$}\\ 

\multicolumn{1}{l|}{w/o cross-split label correction} &
{\color{red}70.20}\tiny$\pm0.16$& \multicolumn{1}{c|}{{\color{red}65.74}\tiny$\pm0.10$} &
{\color{red}{31.77}}\tiny$\pm0.32$& \multicolumn{1}{c|}{{\color{red}{15.93}}\tiny$\pm0.21$} &
82.38\tiny$\pm0.16$& \multicolumn{1}{c|}{82.10\tiny$\pm0.23$} &
{\color{red}69.61}\tiny$\pm0.65$& \multicolumn{1}{c}{{\color{red}{59.67}}\tiny$\pm0.11$}\\

\bottomrule
\end{tabular}
}
\vspace{-2mm}
\end{table*}
\begin{table}[t]
\centering
\caption{Performance (\%) under extreme label noise on CIFAR-100. The table shows the best accuracy and the average accuracy over the last 10 epochs. (*) denotes the results we obtain by re-running their public code.}
\label{tab:severe_100}
\small
{
\renewcommand{\arraystretch}{1.0} 
\setlength\tabcolsep{8pt}
\renewcommand{\tabcolsep}{1.0 mm}

\resizebox{\linewidth}{!}{
\begin{tabular}{c | c c c ccc}
\toprule
\multicolumn{1}{c|}{Noise type} & \multicolumn{6}{c}{Symmetric}\\

\midrule

\multicolumn{1}{c|}{Noise ratio}&   \multicolumn{2}{c}{90\%} & \multicolumn{2}{c}{92\%} & \multicolumn{2}{c}{95\%}\\
\midrule

\multicolumn{1}{c|}{Method}& Best & \multicolumn{1}{c|}{Last} & Best & \multicolumn{1}{c|}{Last} & Best & \multicolumn{1}{c}{Last}\\
\midrule

\multicolumn{1}{c|}{UNICON \cite{karim2022unicon}} & 44.82& \multicolumn{1}{c|}{44.51} & \multicolumn{1}{l}{32.08*}& \multicolumn{1}{l|}{31.85*}& \multicolumn{1}{l}{19.12*}& \multicolumn{1}{l}{18.14*}\\

\multicolumn{1}{c|}{CrossSplit (ours)} & \bf{52.40}& \multicolumn{1}{c|}{\bf 52.05} & \multicolumn{1}{l}{\bf{46.25}}& \multicolumn{1}{l|}{\bf 45.85}& \multicolumn{1}{l}{\bf{29.97}}& \multicolumn{1}{l}{\bf 29.57}\\
\bottomrule
\end{tabular}
}
}

\vspace{-4mm}
\end{table}

\subsection{Additional results under extreme label-noise }
\begin{figure}
  \centering
  \includegraphics[width=1.0\linewidth]{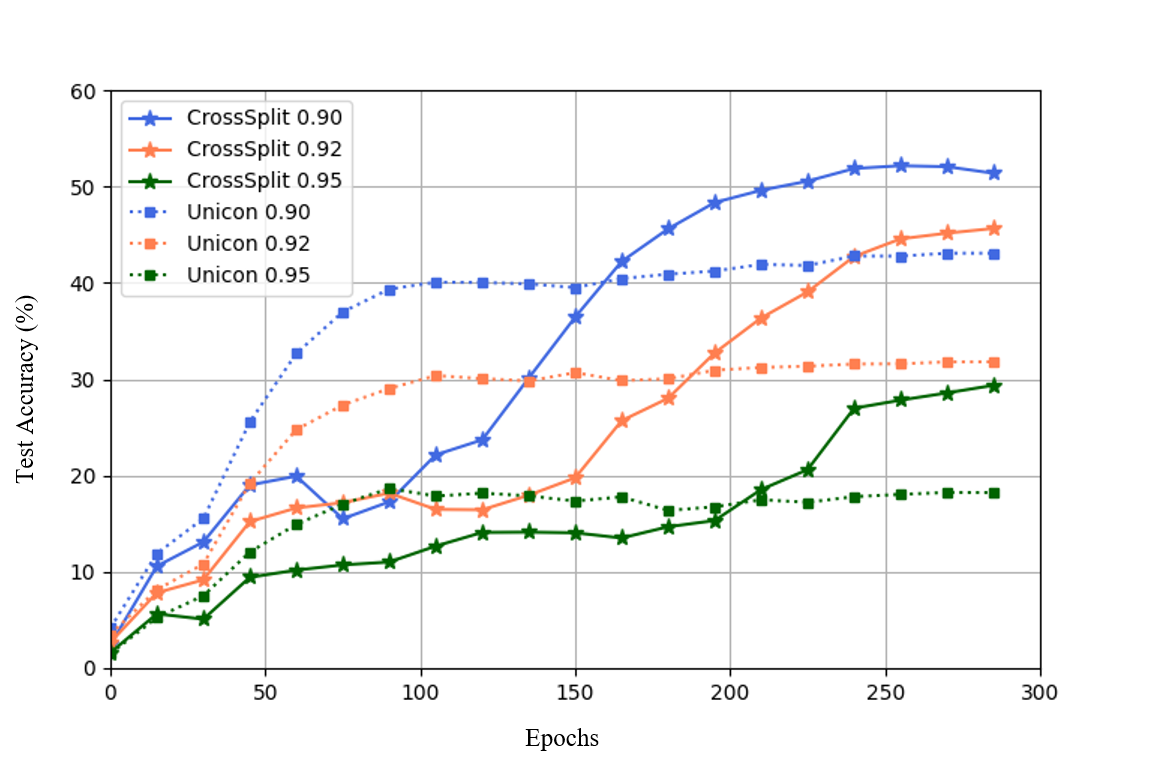}
  \caption{
  Comparisons of test accuracy (\%) of \emph{CrossSplit} and UNICON under extreme label noise on CIFAR-100. While learning progresses slower for \emph{CrossSplit} at the beginning (possibly due to the untrained peer network not effectively correcting labels yet), the final performance is consistently superior to UNICON.
  }
  \label{fig:severe_100}
\end{figure}

\citet{karim2022unicon} show excellent performance of the current state-of-the-art UNICON  even in the  case of extremely high levels of label noise (over 90\%). Here we provide analogous results for \emph{CrossSplit} under extreme noise ratio (90\%, 92\%, and 95\%). 
\cref{tab:severe_100} shows the results for CIFAR-100 with symmetric label noise. 
The performance of UNICON (except for label noise of $90\%$) is obtained by re-running their publicly available code\footnote{\href{https://github.com/nazmul-karim170/UNICON-Noisy-Label}{https://github.com/nazmul-karim170/UNICON-Noisy-Label}}. 
In \cref{fig:severe_100}, at the early training epochs, the performance of \emph{CrossSplit} (star marked solid line) may seem inferior compared to UNICON (square marked dashed line). This can be interpreted as the fact that some noisy labels are likely to be temporarily included during training due to the lack of a selection mechanism. However, as training proceeds, the effect of noisy labels is gradually minimized by our cross-split label correction process, so it can be confirmed that the performance improves rapidly at later training epochs and consistently at all noise levels.
We observe that \emph{CrossSplit} outperforms UNICON for all noise levels on CIFAR-100 (See \cref{tab:severe_100} and \cref{fig:severe_100}.).

\subsection{Memorization analysis} 
\label{sec:results_memo}

The previously-discussed results show that \emph{CrossSplit} compares well with -- and often outperforms -- the competing baselines. This begs the question of the origin of this performance gap. The core hypothesis of the paper is that our method induces an implicit regularization that better prevents the memorization of noisy labels. In this section, we investigate this hypothesis by quantifying this memorization and comparing it with the current state-of-the-art UNICON \cite{karim2022unicon}. 

To do so, we check the training accuracy separately on the clean and noisy samples of CIFAR-10 and CIFAR-100 with different noise types and ratios (symmetric-50\%, 90\% and asymmetric-40\% noise). The results are shown in \cref{fig:memo}. From left to right, the plots show (a) the training accuracy for noisy (mislabelled) samples, (b) the training accuracy for clean samples, and (c) the test accuracy.

 \paragraph{Discussion} During the initial warm-up period where the whole dataset is used for training, we observe that the noisy samples are increasingly memorized, especially on CIFAR-100 (\cref{fig:memo}). Immediately after the warm-up period though, some forgetting often occurs for both methods, i.e., the accuracy on noisy samples tends to decrease. However, in the case of UNICON, memorization rises again within a few epochs. By contrast, \emph{CrossSplit} manifestly continues to mitigate this memorization while maintaining the fit of clean samples (\cref{fig:memo} (b)). This effect seems to correlate with the gain of performance observed in \cref{fig:memo} (c). 
In summary, we find that \emph{CrossSplit} effectively reduces memorization of noisy labels in contrast to UNICON, which explains its superior performance.

\subsection{Ablation Study} 
\label{sec:ablation}

In this section, we perform an  ablation study to demonstrate the effectiveness of some key components of CrossSplit: data splitting, class-balancing coefficient normalization by $\text{JSD}_\text{norm}$ (\cref{eq:norm_of_JSD}), and cross-split label correction.  
We remove each component to quantify its contribution to the overall performance on CIFAR-10/100 with symmetric-50\% and 90\% noise and CIFAR-10/100 with asymmetric-10\% and 40\% noise. \cref{tab:ablation_cifar10} and \cref{tab:ablation_cifar100} show the test accuracy in different ablation settings for CIFAR-10 and CIFAR-100, respectively. We repeat the experiments three times with different seeds for the random initialization of the network parameters and report averages and standard deviations. 

\paragraph{Data splitting is important}
We first study the effect of data splitting by training each network on the whole training dataset (with no split).
If our training framework had no benefit, then we would expect training on the full dataset to be beneficial, since each network simply sees more labeled data.
However, we observe a degradation of the overall performance -- which is more pronounced on CIFAR-100. Specifically, we observe a 38.21\% drop (from 52.40\% to 14.19\%) in the case of a symmetric-90\%-noise and a 4.66\% drop (from 76.78\% to 72.12\%) in the case of an asymmetric-40\%-noise. This is because the larger the noise level, the greater the effect of memorization.
This tells us that data splitting is an important part in reducing memorization, even though each network sees less labeled data.

\paragraph{Class balancing is highly beneficial when noise is high}
Second, to highlight the effect of class-balancing coefficient normalization, we generate soft labels as in \cref{eq:softlabel} and \cref{eq:beta} but without normalizing the JSD. Somewhat surprisingly, this yields a slight performance \emph{increase} in low-label-noise scenarios. 
However, when the noise ratio is large (symmetric-90\%-noise, asymmetric-40\%-noise), we see that it causes a large performance degradation: there is a drop of 15.71\% (from 91.25\% to 75.54\%) for symmetric-90\%-noise on CIFAR-10; and there are 
drops of 19.03\% (from 52.40\% to 33.37\%) for symmetric-90\%-noise and 5.19\% (from 76.78\% to 71.59\%) for asymmetric-40\%-noise on CIFAR-100. 
Moreover, when class-balancing normalization is not used, it can cause divergence in training. 
This yields a big performance gap between the best and the average accuracy, especially in case of high noise scenarios (i.e in \cref{tab:ablation_cifar100}, Best: 71.59\% vs. Last: 60.35\% for asymmetric-40\%-noise). This shows the importance of taking into account the class-wise difficulty, especially in the case of high noise ratio -- as first demonstrated by \citet{karim2022unicon}. 

\paragraph{Cross-split label correction is crucial}
Third, we demonstrate the benefit of cross-split label correction. When we only use the assigned label with no correction, we find a huge performance degradation -- especially on CIFAR-100, which is known \cite{pleiss2020identifying} to contain many more ambiguous examples compared to CIFAR-10.
Especially, when the noise ratio is large (symmetric-90\%-noise, asymmetric-40\%-noise) on CIFAR-100, there are 
drops of 20.63\% (from 52.40\% to 31.77\%) for symmetric-90\%-noise and 7.17\% (from 76.78\% to 69.61\%) for asymmetric-40\%-noise.
This demonstrates the value of our label correction procedure using the peer network.

\section{Conclusion}
\label{sec:concl} 
This paper introduces a new framework for learning with noisy labels, which builds and improves upon existing methods based on label correction and co-teaching techniques. By using a pair of networks trained on two disjoint parts of the labelled dataset, our method bypasses the sample selection procedure used in recent state-of-the-art methods, which can be subject to selection errors. We propose data splitting, cross-split label correction by the peer network prediction, and class-balancing coefficient normalization that is effective in dealing with noisy labels. Our experimental results demonstrate the success of the method at mitigating  the memorization of noisy labels, and show that it achieves state-of-the-art classification performance on several standard noisy benchmark datasets: CIFAR-10, CIFAR-100, and Tiny-ImageNet, with a variety of noise ratios. 
Most importantly, we also demonstrate that our method outperforms the state-of-the-art on the naturally noisy dataset mini-WebVision, which brings our model closer to real world application. We discuss limitations and future work in  \cref{limitation_and_futurework}.

{\small
\bibliographystyle{icml/icml2023}
\bibliography{egbib}
}

\newpage
\appendix
\addcontentsline{toc}{chapter}{APPENDICES}


\section{Implementation Details}
\subsection{Detail on Relaxation Parameter}
\label{relaxation}
As mentioned in \cref{eq:beta} of the main paper, we use a relaxation parameter $\gamma$ as way to control the range of the combination coefficients in our definition of the soft labels. 
In our experiments, $\gamma$ gradually increases from 0.6 to 1 during training according to the following schedule: 
\begin{equation}
\gamma=
\begin{cases}
  0.6, & \text{if}\ epoch \in[E_\text{warm}, E_\text{warm}+2\delta] \\
  0.8, & \text{else if}\ epoch \in[E_\text{warm}+2\delta, E_\text{warm}+3\delta] \\
  1, & \text{otherwise}
\end{cases}
\label{eq:gamma}
\end{equation}
where the parameter $\delta$ determines the relaxation period. We set it to 10. 
\subsection{Training details}
\label{training_details}
The training details are summarized in \cref{tab:traindetail}.
For CIFAR-10 and CIFAR-100, we train each network using stochastic gradient descent (SGD) optimizer with momentum 0.9 and a weight decay of 0.0005.
Training is done for 300 epochs with a batch size of 256. 
We set the initial learning rate as 0.01 and use a a cosine annealing decay \cite{loshchilov2016sgdr}. Just like in \cite{dividemix, karim2022unicon}, a warm-up training on the entire dataset is performed for 10 and 30 epochs for CIFAR-10 and CIFAR-100, respectively. 
For Tiny-ImageNet, we use SGD with momentum 0.9, a weight decay of 0.0005, and a batch size of 40. We train each network for 360 epochs, which includes a warm-up training of 10 epochs.
For mini-WebVision, we use SGD with momentum 0.9, a weight decay of 0.0005, and a batch size of 128. We train the networks for 140 epochs with a warm-up period. We also set the initial learning rate to 0.02 and decay it with decay factor 0.1 with intervals of 80 and 105.

\section{T-SNE Visualization}
\label{sec:tsne}
\begin{figure*}[ht]
  \centering
  \includegraphics[width=\linewidth]{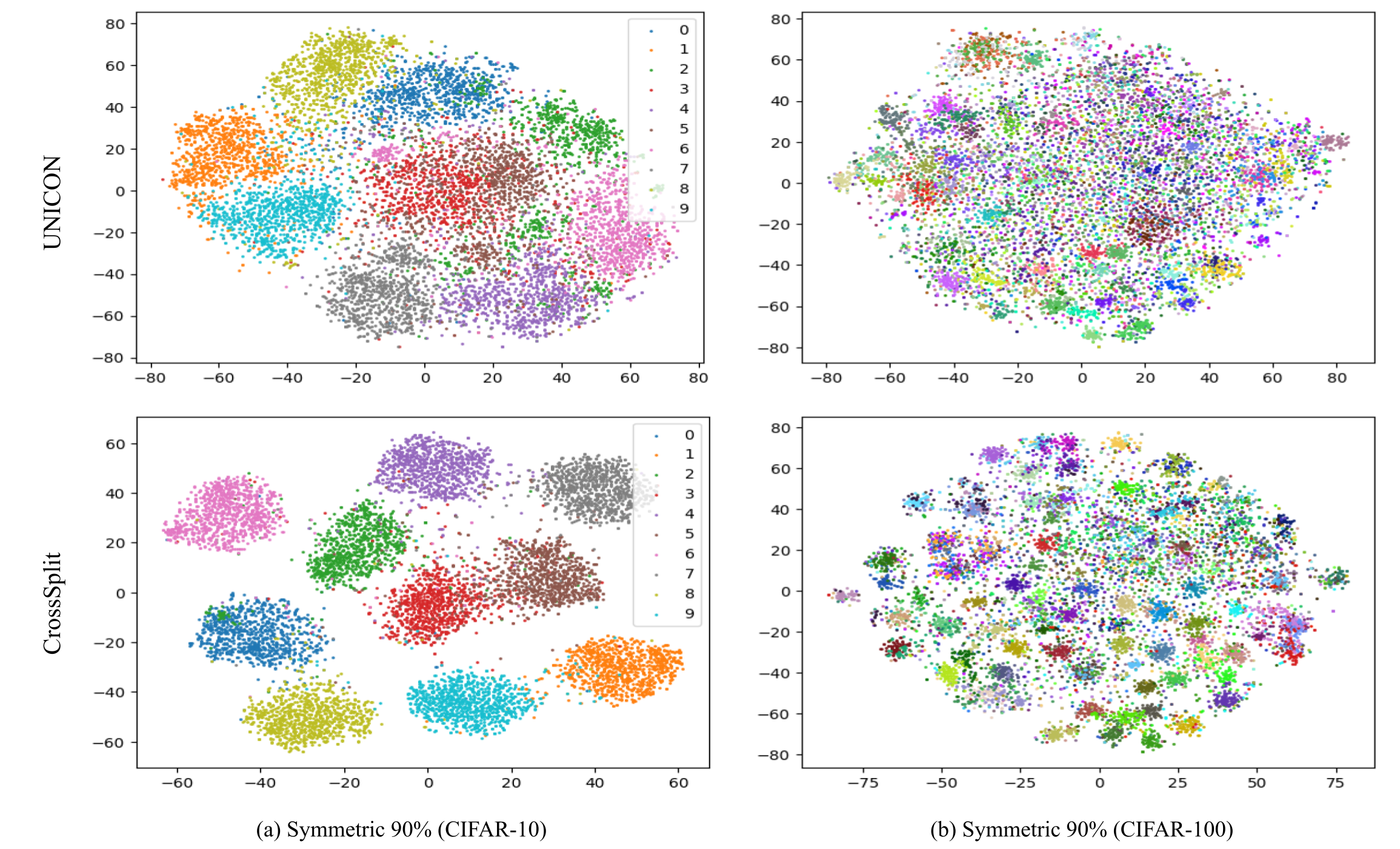}
  \caption{
  T-SNE visualizations of learned features of test images by UNICON \cite{karim2022unicon} and CrossSplit with symmetric noise of 90\%. In general, the clusters for {CrossSplit} are significantly better separated than for UNICON. This is evidence for the superior representation learned by reducing memorization of noisy labels through {CrossSplit}.
  }
  \label{fig:tsne_sym}
\end{figure*}

\begin{figure*}[ht]
  \centering
  \includegraphics[width=\linewidth]{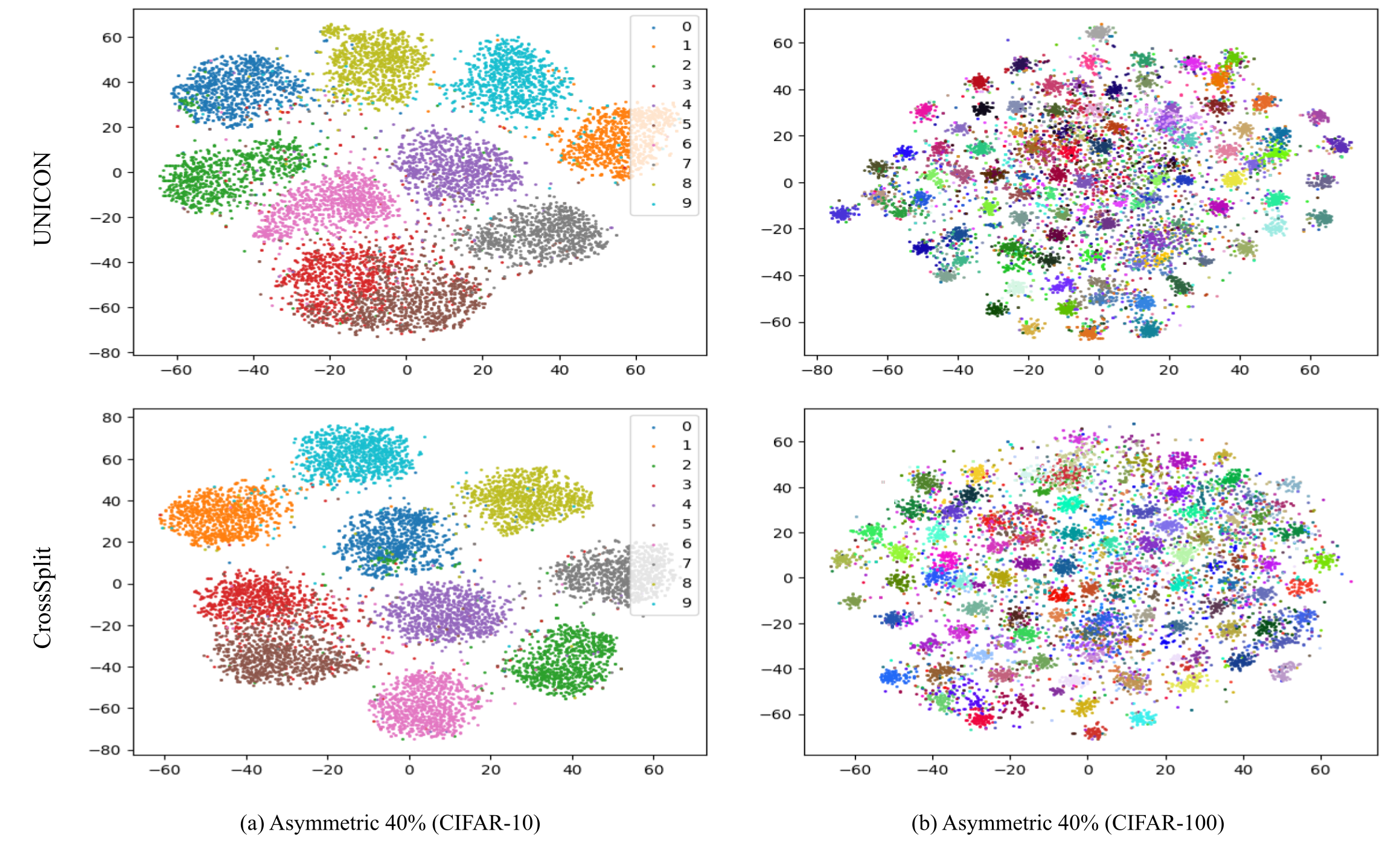}
  \caption{
  T-SNE visualizations of learned features of test images by UNICON \cite{karim2022unicon} and CrossSplit with asymmetric noise of 40\%.
  }
  \label{fig:tsne_asym}
\end{figure*}

In this section, we provide a visual comparison of the features (penultimate layer) learned by UNICON \cite{karim2022unicon} and \emph{CrossSplit}.  \cref{fig:tsne_sym} and \cref{fig:tsne_asym} show the class distribution of the features corresponding to test images on CIFAR-10 and CIFAR-100 with 90\% symmetric and 40\% asymmetric noise, respectively.  This suggests that the representations learned by \emph{CrossSplit} do a better job at separating the classes than UNICON. 

\section{Additional Ablation Results}

\paragraph{Effect of Contrastive Loss} As mentioned in Sec. 2.2, following \cite{karim2022unicon}, we use a contrastive loss $L_\text{con}$ in addition to the  semi-supervised loss for the training of the two networks. Here we show ablation over this unsupervised learning component. 

The results are shown in  \cref{tab:ablation_sym}. We observe that the contrastive loss is particularly helpful in improving the performance in a high noise regime (90\%).
\begin{table}[t]

\centering
\caption{{Test accuracy (\%) for different loss combinations on CIFAR-100 with symmetric label noise.} }
\label{tab:ablation_sym}
\small
{

\renewcommand{\arraystretch}{1.0}
\setlength\tabcolsep{8pt}
\renewcommand{\tabcolsep}{1.0 mm}
\resizebox{0.75\linewidth}{!}{
\begin{tabular}{c| c c |cc}
\toprule

\multicolumn{1}{c|}{Dataset} &\multicolumn{4}{c}{CIFAR-100} \\

\midrule
\multicolumn{1}{c|}{Noise type} &\multicolumn{4}{c}{Symmetric} \\

\multicolumn{1}{c|}{Noise ratio} & \multicolumn{2}{c|}{$50\%$}& \multicolumn{2}{c}{$90\%$}   \\

\midrule

\multicolumn{1}{c|}{Method} & \multicolumn{1}{c}{Best} & \multicolumn{1}{c|}{Last}& \multicolumn{1}{c}{Best} & \multicolumn{1}{c}{Last} \\

\midrule

\multicolumn{1}{l|}{CrossSplit} & \bf{75.72} & 75.50 & \bf{52.40} & \bf{52.05}  \\ 

\multicolumn{1}{l|}{CrossSplit w/o $L_\text{con}$}&75.68&\bf{75.58} & 31.42 & 31.15 \\ 

\bottomrule
\end{tabular}
}
}

\end{table}
\begin{table}[h]
\centering
\caption{{Training details on CIFAR-10, CIFAR-100, Tiny-ImageNet and Mini-WebVision datasets.}}
\label{tab:traindetail}
\small
{
\renewcommand{\arraystretch}{1.5}
\setlength\tabcolsep{8pt}
\renewcommand{\tabcolsep}{1.2 mm}
\resizebox{\linewidth}{!}{
\begin{tabular}{c | c c c c}
\toprule
\multicolumn{1}{c|}{Dataset} & CIFAR-10 & CIFAR-100 & Tiny-ImageNet & mini-WebVision\\

\midrule
\multicolumn{1}{c|}{Batch size}& 256 & 256 & 40 & 128 \\
\multicolumn{1}{c|}{Network}& PRN-18 &PRN-18&PRN-18& ResNet-18 \\
\multicolumn{1}{c|}{Epochs}& 300 & 300 & 360 & 140 \\
\multicolumn{1}{c|}{Optimizer}& SGD & SGD & SGD & SGD\\
\multicolumn{1}{c|}{Momentum}& 0.9 & 0.9& 0.9 & 0.9\\
\multicolumn{1}{c|}{Weight decay}& 5e-4 & 5e-4& 5e-4 & 5e-4\\
\multicolumn{1}{c|}{Initial LR}& 0.1 & 0.1 & 0.005 & 0.02 \\
\multicolumn{1}{c|}{LR scheduler}& \multicolumn{3}{c}{Cosine Annealing LR} & \multicolumn{1}{c}{Multi-Step LR} \\
\multicolumn{1}{c|}{$T_\text{max}$/\text{LR decay factor}}& 300 & 300 & 360 & 0.1 (80, 105) \\
\multicolumn{1}{c|}{Warm-up period}& 10 & 30 & 10 & 1  \\
\bottomrule
\end{tabular}
}
}

\end{table}

\begin{table}[t]
\centering
\caption{Performance (\%) under extreme label noise on CIFAR-10. The baseline results are imported from \cite{karim2022unicon}.}
\label{tab:severe_10}
\small
{
\renewcommand{\arraystretch}{1.0} 
\setlength\tabcolsep{8pt}
\renewcommand{\tabcolsep}{1.0 mm}

\resizebox{\linewidth}{!}{
\begin{tabular}{c | c c c ccc}
\toprule
\multicolumn{1}{c|}{Noise type} & \multicolumn{6}{c}{Symmetric}\\

\midrule

\multicolumn{1}{c|}{Noise ratio}&   \multicolumn{2}{c}{90\%} & \multicolumn{2}{c}{92\%} & \multicolumn{2}{c}{95\%}\\
\midrule

\multicolumn{1}{c|}{Method}& Best & \multicolumn{1}{c|}{Last} & Best & \multicolumn{1}{c|}{Last} & Best & \multicolumn{1}{c}{Last}\\
\midrule

\multicolumn{1}{c|}{DivideMix \cite{dividemix}}& 76.08 & \multicolumn{1}{c|}{-} & 57.62 & \multicolumn{1}{c|}{-} & {51.28} & \multicolumn{1}{c}{-} \\
\multicolumn{1}{c|}{UNICON \cite{karim2022unicon}}& 90.81& \multicolumn{1}{c|}{89.95} & \bf{87.61}& \multicolumn{1}{c|}{-} & \bf{80.82}& \multicolumn{1}{c}{-} \\

\multicolumn{1}{c|}{CrossSplit (ours)}&  \bf{91.25}& \multicolumn{1}{c|}{\bf 91.02} & \underline{84.45} & \multicolumn{1}{c|}{84.07}& \underline{62.73}& \multicolumn{1}{c}{62.42} \\
\bottomrule
\end{tabular}
}
}

\vspace{-4mm}
\end{table}

\section{Limitation and Future Work}
\label{limitation_and_futurework}
Our work shows that data splitting and cross-split training techniques can boost the robustness of deep learning models under label noise in a wide range of noise ratios. However, this was not the case in {\it all} situations we considered: we thus observed a degradation of performance for Tiny-ImageNet with 50$\%$ symmetric noise in \cref{tab:tiny}, as well as for CIFAR10 under extreme noise ratios (over 92\%, see \cref{tab:severe_10}). Even though we should of course not expect any {\it free lunch}, i.e. universal improvement across all situations, we believe the analysis of such a negative result and its dependence on the dataset and noise ratio would be a way to better understand the reasons and conditions of success of our method.  

We restricted our study to image classification tasks in this paper. This is also the stage of many prior works in this line of research. However, we expect learning with noisy labels to come with its own challenges in other domains such as text classification \cite{zhu2022bertlabel}. Extending our study to this domain is an interesting avenue for future work.

\end{document}